# Syllable analysis to build a dictation system in Telugu language

*N.Kalyani*
*Assoc. Prof*, CSE Dept
*G.N.I.T.S*
*Hyderabad, India.*

*Dr K.V.N.Sunitha*
*.Professor & HOD, CSE Dept.*
*G.N.I.T.S*
*Hyderabad, India.*

*Abstract*— **In recent decades, Speech interactive systems gained increasing importance. To develop Dictation System like Dragon for Indian languages it is most important to adapt the system to a speaker with minimum training. In this paper we focus on the importance of creating speech database at syllable units and identifying minimum text to be considered while training any speech recognition system. There are systems developed for continuous speech recognition in English and in few Indian languages like Hindi and Tamil. This paper gives the statistical details of syllables in Telugu and its use in minimizing the search space during recognition of speech. The minimum words that cover maximum syllables are identified. This words list can be used for preparing a small text which can be used for collecting speech sample while training the dictation system. The results are plotted for frequency of syllables and the number of syllables in each word. This approach is applied on the CIIL Mysore text corpus which is of 3 million words.**

## I. INTRODUCTION *(HEADING 1)*

Developing a robust dictation system to transcribe continuous speech signal into a sequence of words is a difficult task, as continuous speech does not have any natural pauses in between words. It is also difficult to make the system robust for speaker variability and the environment conditions. There are many research organizations working on speech with different approach. The conventional method of building a large vocabulary speech recognizer for any language uses a top-down approach to speech recognition (Huang & Acero 1993)[1]. What they mean by top-down is that these recognizers first hypothesize the sentence, then the words that make up the sentence and ultimately the sub-word units that make up the words. This approach requires large speech corpus with sentence or phoneme level transcription of the speech utterances (Thomas Hain *et al* 2005; Ravishankar 1996) [2]. The transcriptions must also include different speech order so that the recognizer can build models for all the sounds present. It also requires maintaining a dictionary with the phoneme/sub-word unit transcription of the words and language models to perform large vocabulary continuous speech recognition. The recognizer outputs words that exist in the dictionary. If the system is to be developed for a new language it requires building of a dictionary and extensive language models for the new language. In country like India which includes 22 officials and a number of unofficial languages, building huge text and speech databases is a difficult task. There are other related works which gained good importance is listed below.

Some methods that require manually annotated speech corpora for speech recognition are listed. A method called bootstrapping is proposed by Rabiner *et al* (1982)[3] which can increase the transcribed data for training the system, for speech recognition. Ljolje & Riley (1991)[4] have used an automatic approach to segmentation and labeling of speech when only the orthographic transcription of speech is available. Kemp&Waibel (1998)[5] used unsupervised training approach for speech recognition for TV broadcasts. Wessel&Ney (2001)[6] have proposed an approach in which a low-cost recognizer trained with one hour of manually transcribed speech is used to recognize 72 hours of unrestricted acoustic data. Lamel *et al* (2002)[7] have shown that the acoustic models can be initialized using as little as 10 minutes of manually annotated data.

There are also few methods that do not require any manually annotated speech corpora for speech recognition. Incremental maximum a posteriori estimation of HMMs is proposed by Gotoh & Hochberg (1995)[8]. This algorithm randomly selects a sub-set of data from the training set, updates the model using maximum a posteriori estimation and this process is iterated until it covers all possible units. Chang *et al* (2000)[9] proposed a method which extracts articulatory-acoustic phonetic features from each frame of speech signal and then the phone is identified using neural networks. There is an interesting approach proposed by Nagarajan & Murthy (2004)[10] for Indian languages. Their approach focuses on automatically segmenting and transcribing the continuous speech signal into syllable-like units using a group delay based segmentation algorithm without the use of manually segmented and labeled speech corpora. This approach is more appropriate for Indian languages as they are syllable centered.

The focus of this paper is to extract the possible syllables from the raw Telugu text corpus. Once the syllable like units is obtained they are analyzed to understand their frequency of coverage in words. The final list of words can be prepared considering the words having high probable syllables or identifying the words which contain the maximum syllables.

Details of Telugu corpus and Procedure to convert Telugu text to WX notation is introduced in section 2. Algorithm which uses Syllabification rules to syllabify the text is discussed in section 3. A study of results obtained is summarized in section 4 and conclusion and future scope presented in section 5.

## II. CONVERSION OF TELUGU TEXT TO WX NOTATION.

Telugu is one of the major Scheduled languages of India. It has the second largest number of speakers mainly concentrated in South India. It is the official language of Andhra Pradesh and second widely spoken language in Tamilnadu, Karnataka. There are number of Telugu language speakers have migrated to Mauritius, South Africa, and recently to USA, UK, and Australia. Telugu is often referred as "Italian of the East".

The Primary units of Telugu alphabet are syllables, therefore it should be rightly called a syllabic language. There is good correspondence in the written and spoken form of the south Indian languages. Any analysis done on written form would closely relate to spoken form of the language.

The Telugu alphabet can be viewed as consisting of more commonly used inventory, a common core, and an overall pattern comprising all those symbols that are used in all domains. The overall pattern consists of 60 symbols, of which 16 are vowels, 3 vowel modifiers, and 41 consonants.

Since Indian languages are syllable-timed languages, syllable is considered as the basic unit in this work and analysis is performed to identify the words with syllables with high frequency and words with varying coverage of syllables.

### A. *Telugu to English letter translation*

The WX notation of thirteen vowel signs అ ’ఆ ’ఇ’ఈ ’ఉ’ఊ ’ఋ ’ఎ’ఏ’ఐ ’ఒ’ఓ’ఔ is a,A,i,I,u, U,q,eV,e,E,oV,o,O occur as stand alone characters and In UNICODE Standard 3.0., each of these is assigned a hexadecimal code point 0C00-0C7F. When a vowel occurs immediately after a consonant it is represented as a dependent or secondary sign called, guNiMtaM gurtulu. The Telugu alphabet is a syllabic language in which the primary consonant always has an inherent vowel [a] / ✓ /. When a consonant is attached with another vowel other than [a] / ✓ / then secondary vowel sign is attached to the consonant after removing the inherent vowel /a/. There are exceptions where the primary vowel may be considered as secondary.

There are 41 consonants in the common core inventory. In Unicode Standard 3.0 they begin with 0C15 to 0C39 and 0C1A to 0C2F. The character set for consonants in Telugu is complex and peculiar in their function. These character signs have three or more than three distinct shapes depending on their occurrence

- Base consonants or Primaries, when they are used as stand alone characters.
- Pure consonant or hanger, when used with a vowel other than the inherent vowel /a/
- Ottulu or Secondary consonant, when used as a constituent of a conjunct

The basic character set for consonants are called as primaries or stand alone characters as they occur in the alphabet. Each of which has an inherent vowel /a/ which often is explicitly indicated by sign / ✓ /. This graphic sign indicating the vowel /a/ is normally deleted and replaced with another explicit mark for a different vowel.

List of pure consonants carrying explicit secondary vowel /a/ sign and its corresponding WX notation are క-ka గ-ga ఖ-Ka ఘ-G ఙ-fa , చ-ca ఛ-Ca జ-ja ఝ-Ja ఞ-Fa , ట-ta ఠ-Ta డ-da ఢ-Da ణ-Na , త-wa థ-Wa ద-xa ధ-Xa న-na , ప-pa ఫ-Pa బ-ba భ-Ba మ-ma య-ya ర-ra ల-la వ-va శ-Sa ష-Ra స-sa హ-ha ళ-lYa ఱ-rY

The Telugu text in Unicode format is converted to WX notation. The conversion is done character by character using Unicode value of the character. If the Unicode of the character is between 0C15 and 0C39 (క to హ), representation corresponding to Pure consonant is retrieved from WX table. If the Unicode of the character is between 0C3E and 0C4C (ా to ౌ), the last letter from Pure consonant is removed and secondary vowel representation is added. If Unicode of the character is 0C4D which correspond to stress mark ్, the last letter from the WX notation is removed indicating that the next occurrence of character is secondary consonant.

### B. *Algorithm*

The algorithm for conversion is given below where englishtext is initialized to null.

- string englishtext=null
- read the contents and convert into character array
  - ***for*** each character till end of the file do
    - ***if*** Unicode of the letter is between 0C15 and 0C60

      retrieve the corresponding English character for the Unicode, add to englishtext and increment i by 1

- ***else if*** Unicode of the letter is between 0C3E and 0C4C

  remove the last letter from the englishtext, retrieve the corresponding English character for the Unicode, add to englishtext and increment i by 1

- ***else if*** Unicode of the letter is 0C4D

  remove last letter from the englishtext

- ***else***

  copy the character into English text and increment i by 1

o ***end for***

- ***Store*** in temp file for Syllabification.
- ***end***

## III. SYLLABIFICATION

The scripts of Indian languages have originated from the ancient Brahmi script. The basic speech sounds units and basic written form has one to one correspondence. An important feature of Indian language scripts is their phonetic nature. The characters are the orthographic representation of speech sounds. A character in Indian language scripts is close to syllable and can be typically of the following form: C, V, CV, CCV and CVC, where C is a consonant and V is a vowel. There are about 35 consonants and about 15 vowels in Indian languages. The rules required to map the letters to sounds of Indian languages are almost straight forward. All Indian language scripts have common phonetic base.

The majority of the speech recognition systems in existence today use an observation space based on a temporal sequence of frames containing short-term spectral information. While these systems have been successful [10, 12], they rely on the incorrect assumption of statistical conditional independence between frames. These systems ignore the segment-level correlations that exist in the speech signal. The high-energy regions in the Short Term Energy function correspond to the syllable nuclei while the valleys at both ends of the nuclei determine the syllable boundaries.

The text segmentation is based on the linguistic rules derived from the language. Any syllable based language can be syllabified using these generic rules. To make the text segments exactly equivalent to the speech units.

The syllable can be defined as a vowel nucleus supported by consonants on either side, It can be generalized as a C*VC* unit where C is a consonant and V is a vowel. The linguistic rules to extract the syllables segments from a text are generated from spoken Telugu. These rules can be generalized to any syllable centric language. The text is preprocessed to remove any punctuation. The following algorithm divides the word into syllable like units.

### A. *Algorithm*

- Read from the file which has text in WX notation.
- Label the characters as consonants and vowels using the following rules
  - o Any consonant except(y, H, M) followed by y is a single consonant, label it as C
  - o Any consonant except (y, r, l, lY, lYY) followed by r is taken as single consonant
  - o Consonants like(k, c, t, w, p, g, j, d, x, b, m, R, S, s) followed by l is taken as single consonant.
  - o Consonant like (k, c, t, w, p, g, j, d, x, b, R, S, s, r) followed by v is taken as a single consonant.
  - o Label the remaining as Vowel (V) or Consonant(C) depending on the set to which it belongs.
  - o Store the attribute of the word in terms of (C*VC*)* in temp2 file.
- For each word in the corpus get its label attribute from temp2 file.
  - o If the first character is a C then the associate it to the nearest Vowel on the right.
  - o If the last character is a C then associate it to the nearest Vowel on the left.
  - o If sequences correspond to VV then break is as V-V.
  - o Else If sequence correspond to VCV then break it as V-CV.
  - o Else If sequence correspond to VCCV then break it as VC-CV.
  - o Else If sequence correspond to VCCCV then break it as VC-CCV.
  - o The strings separated by – are identified as syllable units.
- repeat
- Store the result in output file.

The following Table:1 shows the output obtained for the input in Telugu text in UNICODE

TABLE I. OUTPUT FOR ALGORITHM 1 AND ALGORITHM 2

| S. No | Input | Output of Algoritm 1 | Output of Algorithm 2 |
|---|---|---|---|
| 1. | కంపెనీకంటె | kaMpeVnIkaMteV | kaM-peV-nI-kaM-teV |
| 2. | ఖర్చుకంటె | KarcukaMteV | Kar-cu-kaM-teV |
| 3. | లాభాలకు | lABAlaku | lA-BA-la-ku |

## IV. Statistical Alalysis

### A. *Phoneme Analysis:*

The following observations are based on a study made on CIIL Mysore Telugu text corpus of 3 million words of running texts in Telugu. This corpus is first cleaned and words are extracted. The Word frequencies are dropped in order to avoid their skewing effect on the results of character frequencies. These words are broken into syllables using the rules of the language and analysis is performed to study the distribution of phonemes and syllables.

### B. *Phoneme Frequency chart:*

On observing it is found that of Vowels cover nearly 44.98% in total text corpus. It clearly shows that the vowels are the major units in the speech utterance. The vowel modifiers coverage is 3.82% and the consonant coverage is 51.21%. The following Fig 1 gives the details of the analysis.

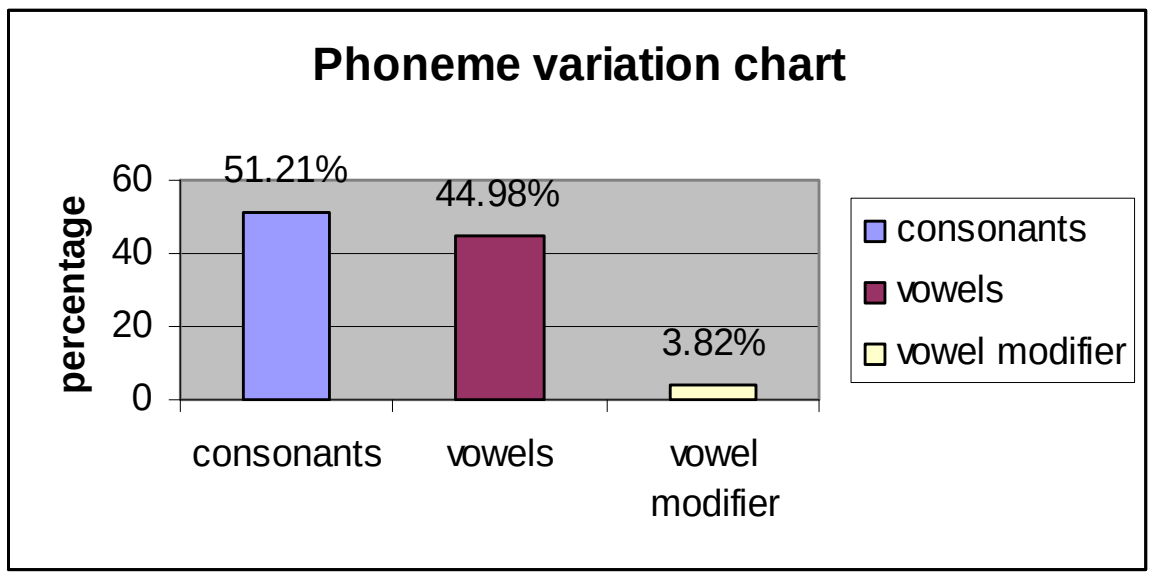

Fig. 1 Phoneme variation chart

### C. *Vowel variation chart:*

Vowels occur either in stand alone form or as modifiers and have total coverage of 48%. The vowels are classified based on the position of the articulator and manner of articulation. The vowel classification is given in Table 5 and the distribution of vowels is shown in the Fig 2.

TABLE 5 Vowel Classification

| **Classification** | **Vowels** |
|---|---|
| Closed Front (CF) | I , i |
| Half Closed Front (HCF) | eV, e |
| Closed Back (CB) | u,U |
| Half Closed Back (HCB) | oV, o |
| Open | a, A |

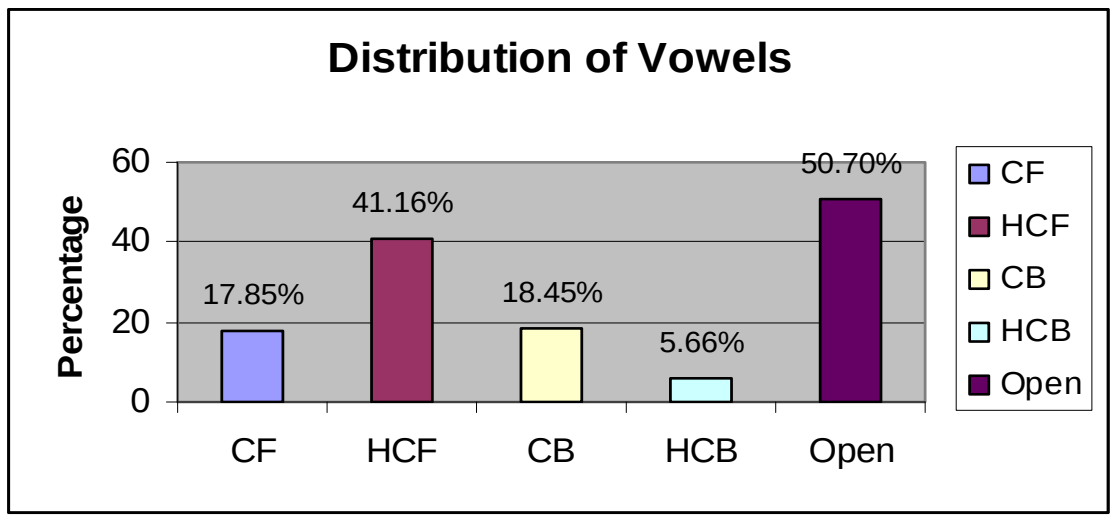

Fig. 2 Vowel distribution chart

### D. *Consonant variation chart:*

Consonants are characterized by significant constriction or obstruction in the pharyngeal and/or oral cavities. Some consonants are voiced and others are unvoiced. Many consonants occur in pairs, that is, they share the same configuration of articulators, and one member of the pair additionally has voicing which the other lacks. Based on the articulators involved and manner of articulation the consonants are classified as Bilabial, Dental Alveolar, Retroflex, Velar and Glottal. The distribution of consonants is shown in the following Fig. 3.

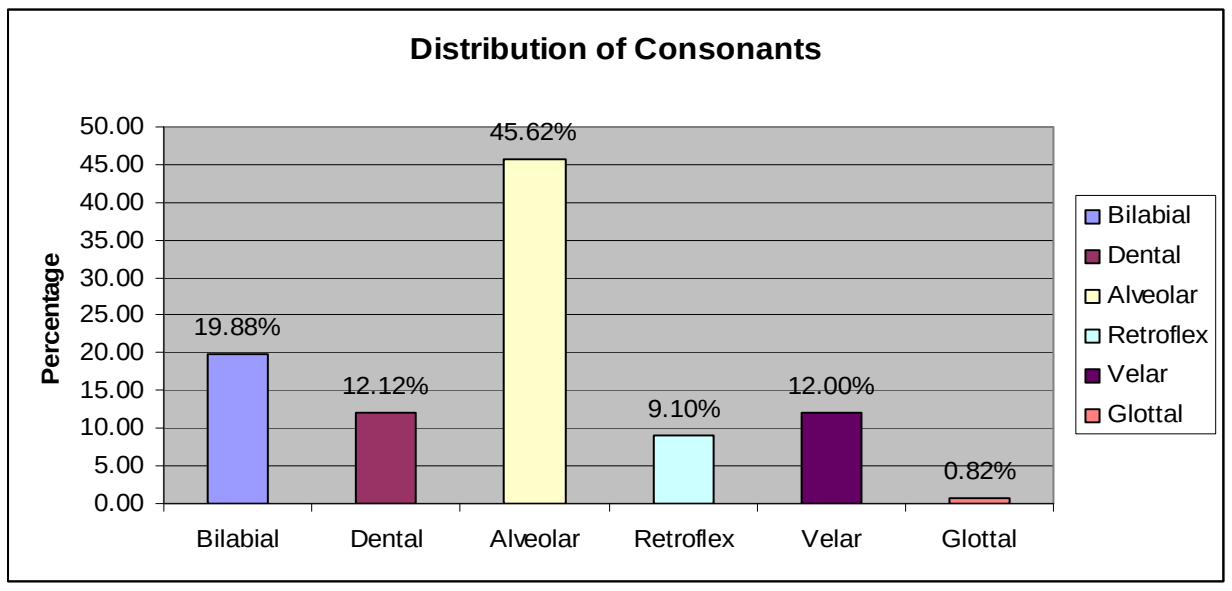

Fig. 3 Consonant distribution chart

Based on the above analysis it is clear that the vowels play a major role in the utterance of speech units. Or the basic unit of utterance is concentric at the vowel which forms the key component in the syllable. It is hence good to build the speech recognition systems considering the syllable as the basic unit.

### E. *Syllable Analysis :*

It is shown that most of the Indian languages are syllable centric. Syllable boundary in speech signal can be approximately identified it is intended to make a study of developing a speech recognition system at Syllable level.

The total distinct syllables observed are 12,378 and the frequency of occurrence of the syllables is plotted in the following chart. The number of Syllables with frequency less than 100 is 11057(12,378 – 1321). It is observed that nearly 4903 syllables have frequency one. This is due to loanwords from English like (Apple, coffee, strength etc.) When these words are written in Telugu it normally takes the same pronunciation. Such kind of words creates a different syllable which may not occur in the native language. Fig 4 shows the count of Syllables with the frequencies in Hundreds.

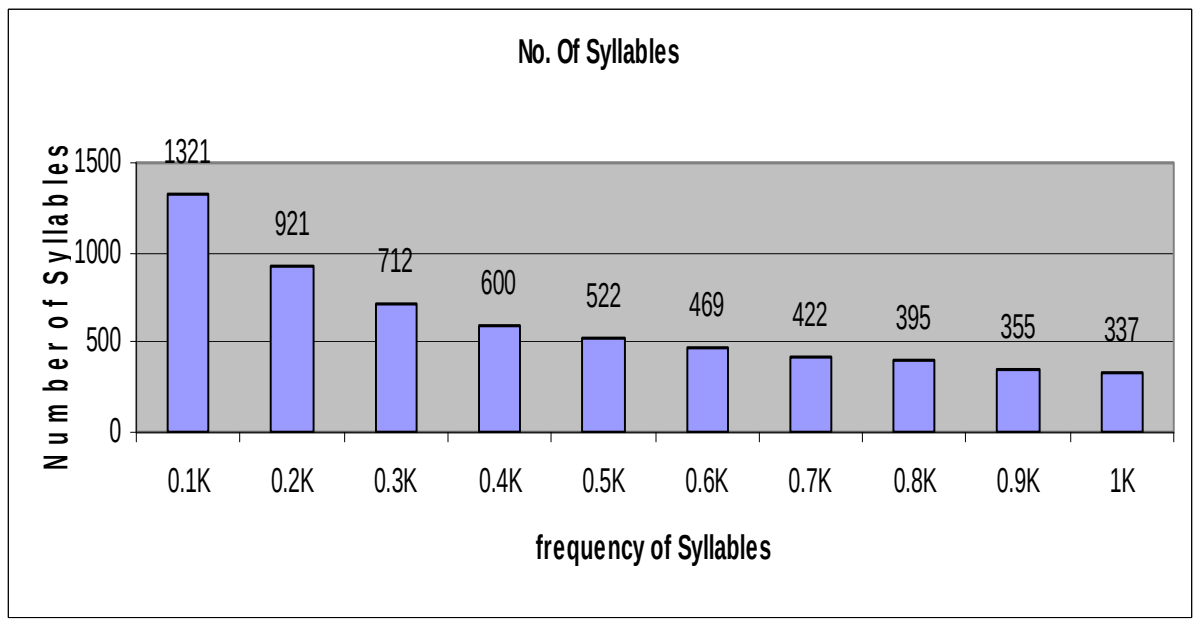

Fig. 4 Number of syllables with frequency in the range 100 to 1K.

The following Fig 5 shows the count of Syllables that have frequency ranging from 1K to 10K.

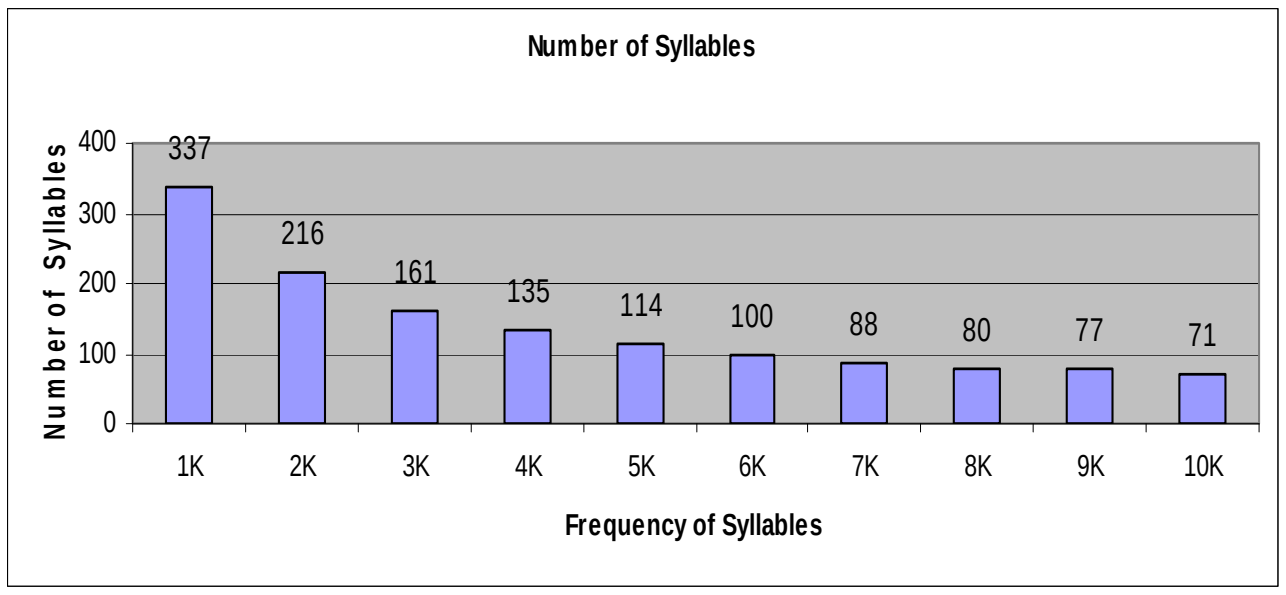

Fig. 5 Number of syllables with frequency in the range 1K to 10K

The following Fig 6 shows the count of Syllables that have frequency ranging from 10K to 100K.

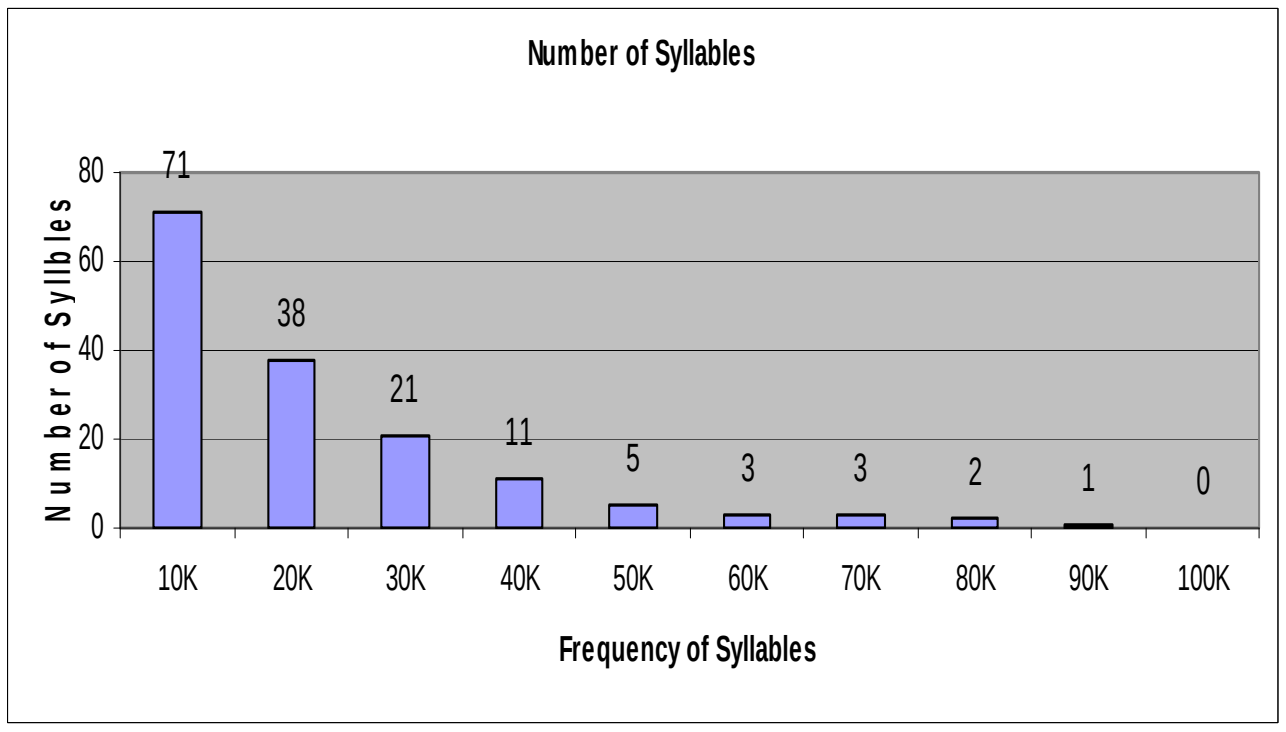

Fig. 6 Number of syllables with frequency in the range 10K to 100K

It is observed that there are nearly 71 syllables that have frequency more than 10K.

A study is also made in terms of the words which have varying number of syllables with varying frequencies. Here in the following figures, plots are given for words which have syllables with cut-off frequency specified on X axis, Y-axis indicates number of words having the Syllable Index and above cut-off frequency and Syllable Index 0.5, 0.8 and 1.0..

Words count, with Syllable Index 50%, 80% and 100% and cut-off frequency varying in the range from 100 to 1000 is shown in Fig.7.

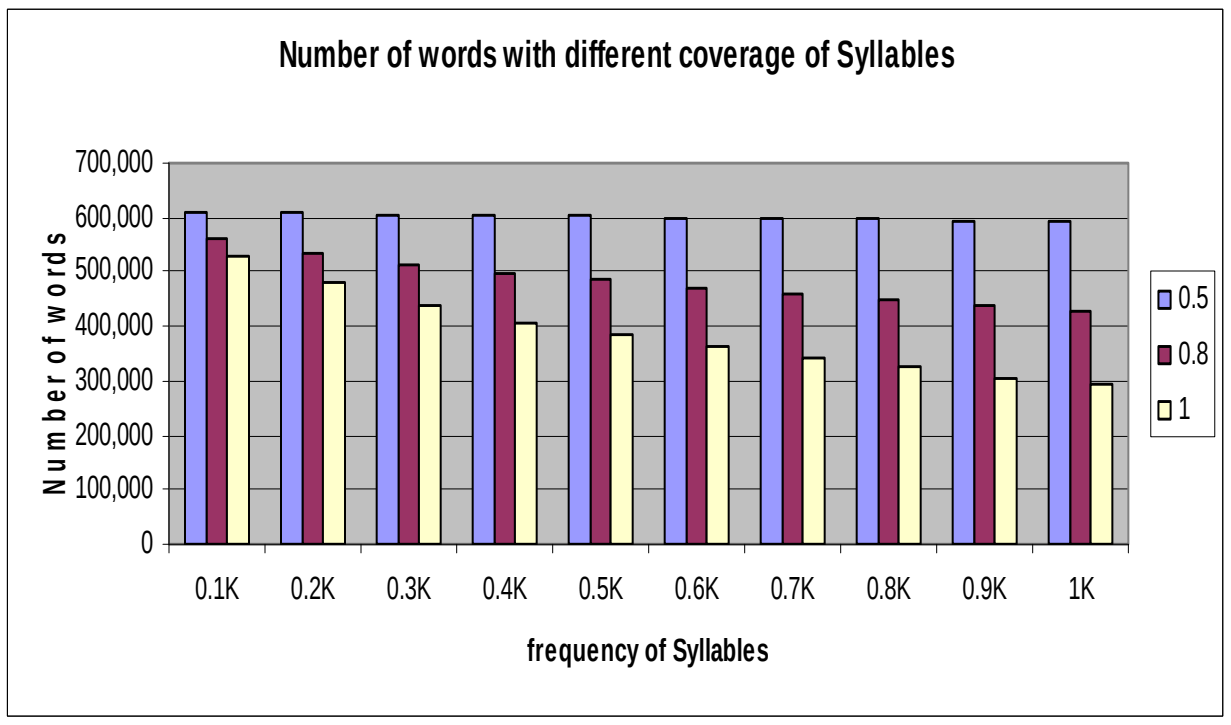

Fig. 7 Number of words having 50%, 80% and 100% syllables with syllable frequency in the range 100 to 1000.

Words count, with Syllable Index 50%, 80% and 100% and cut-off frequency varying in the range from 1K to 10K is shown in Fig.8.

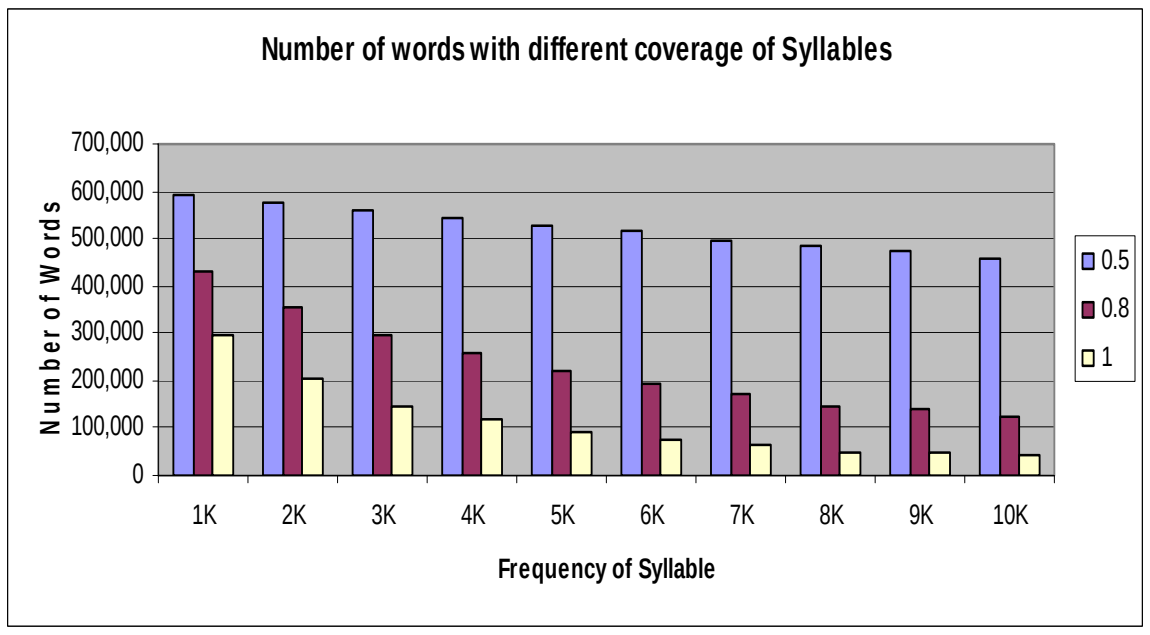

Fig. 8 Number of words having 50%, 80% and 100% syllables with syllable frequency in the range 1K to 10K.

Words count, with Syllable Index 50%, 80% and 100% and cut-off frequency varying in the range from 10K to 100K is shown in Fig.9.

It is observed from the above figures that as the frequency increases the number of words included decreases. Importance of the word depends on the Syllable Index and on the cut-of frequency. It is directly proportional to Syllable Index and cut-of frequency.

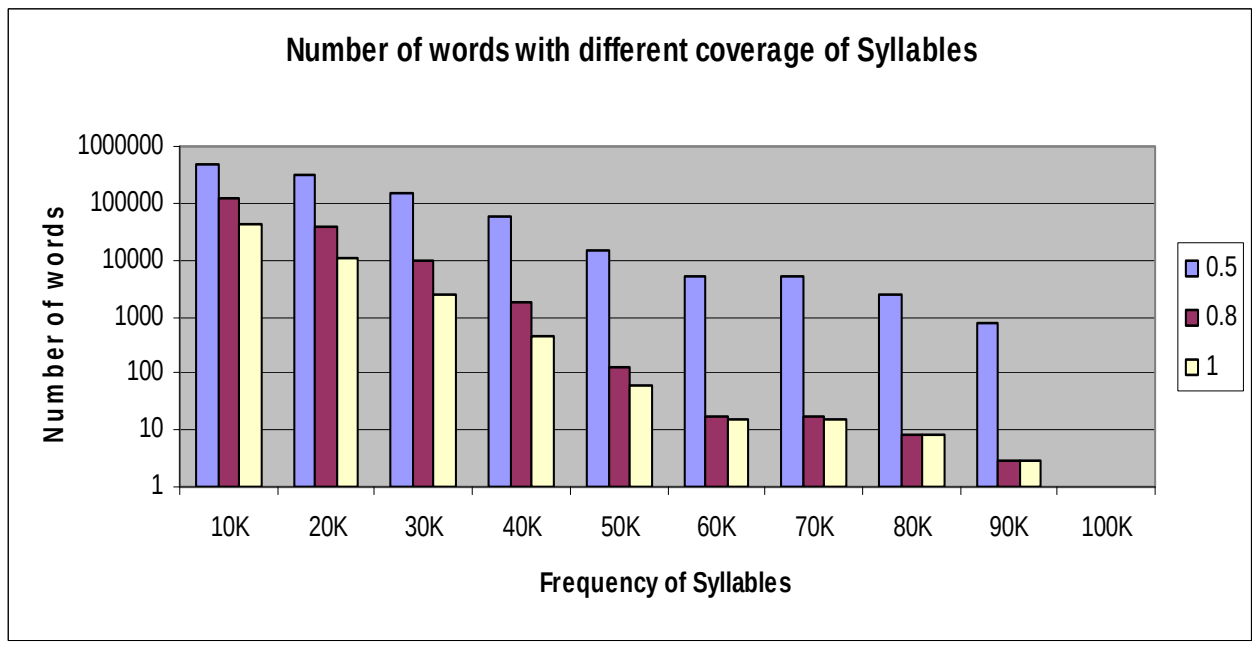

Fig. 9 Number of words having 50%, 80% and 100% syllables with syllable frequency in the range 10K to 100K.

## V. Conclusion

This paper explores the details of phonemes and syllables in the text corpus. As there is one to one correspondence between the written form and spoken form of the language, detailed analysis is performed to understand the coverage of different phonemes and syllables. This analysis is useful in selecting good set of words that would cover all possible syllables in large vocabulary. Optimal selection of words depends on the selection strategy applied. Good strategy can be used to obtain limited words, which are useful during recording the speech to train the system. This ultimately improves the performance of the dictation system.

## AUTHORS PROFILE

**Dr K.V.N.Sunitha** did her B.Tech ECE from Nagarjuna University in 1988, M.Tech Computer Science from REC Warangal in 1993 and Ph.D from JNTU hyderabada in 2006. She has 18 years of Teaching Experience. She has been working in G.Narayanamma Institute of Technology and Science, Hyderabad as HOD CSE Dept. from the inception of the CSE Dept since 2001. She is a recipient of “Academic Excellence Award” by GNITS in 2004.She is awarded “Best computer engineering Teacher Award “ by ISTE in Second Annual Convention held in Feb 2008.She has published more than 35 papers in International & National Journals and Conferences. Authored a book on ”Programming in UNIX and Compiler Design”. She is guiding five PhDs . She is a reviewer for International Journals-JATIT,IJFCA. She is fellow of Institute of engineers &Sr member for IACSIT, International association of CSIT, life member of many technical associations like CSI, IEEE, ACM. She is academic advisory body member for ACE Engg college , Ghatkesar. She is Board of Studies member for UG ang PG programmaes, CSE at G.Pulla Reddy Engg College,Kurnool.

**N. Kalyani** completed B.Tech civil from Osmania University in 1994, M.Tech Computer Science from JNTUH in 2001. She has working experience of 5 years as Design Engineer in R. K. Engineers, Hyderabad and 9 years of teaching for both under and post graduate students. She is presently working as Associate Professor in the Department of Computer Science Engineering, G.Narayanamma Institute of Technology and Science, Hyderabad. She is reciepient of “Academic Excellence Award” by GNITS in 2008. She has published 10 papers in International & National Journals and conferences. She is a life member of CSI & ISTE technical associations.